\documentclass[10pt,twocolumn,letterpaper]{article}
\usepackage{multirow}
\usepackage[preprint]{iccv}      

\usepackage{bm}
\usepackage{mathtools}

\definecolor{iccvblue}{rgb}{0.21,0.49,0.74}
\usepackage[pagebackref,breaklinks,colorlinks,allcolors=iccvblue]{hyperref}

\def\paperID{2851} 
\def\confName{ICCV}
\def\confYear{2025}

\title{\textit{QuoTA}: \underline{Qu}ery-\underline{o}riented \underline{T}oken \underline{A}ssignment via CoT Query Decouple \\ for Long Video Comprehension}

\author{
    Yongdong Luo$^1$, Wang Chen$^1$, Xiawu Zheng$^1$, Weizhong Huang$^1$, Shukang Yin, \\
    Haojia Lin$^1$, Chaoyou Fu$^2$, Jinfa Huang$^3$, Jiayi Ji$^1$, Jiebo Luo$^3$, Rongrong Ji$^1$\\
    $^1$Xiamen University \quad $^2$Nanjing University \quad $^3$University of Rochester
}

\begin{document}
\maketitle
\begin{abstract}

Recent advances in long video understanding typically mitigate visual redundancy through visual token pruning based on attention distribution.
However, while existing methods employ \textbf{post-hoc} low-response token pruning in decoder layers, they overlook the input-level semantic correlation between visual tokens and instructions (query).
In this paper, we propose QuoTA, an \textbf{ante-hoc} training-free modular that extends existing large video-language models (LVLMs) for visual token assignment based on query-oriented frame-level importance assessment. The query-oriented token selection is crucial as it aligns visual processing with task-specific requirements, optimizing token budget utilization while preserving semantically relevant content.
Specifically, 
(i) QuoTA strategically allocates frame-level importance scores based on query relevance, enabling one-time visual token assignment before cross-modal interactions in decoder layers, 
(ii) we decouple the query through Chain-of-Thoughts reasoning to facilitate more precise LVLM-based frame importance scoring,
and (iii) QuoTA offers a plug-and-play functionality that extends to existing LVLMs.
Extensive experimental results demonstrate that implementing QuoTA with LLaVA-Video-7B yields an average performance improvement of \textbf{3.2\%} across six benchmarks (including Video-MME and MLVU) while operating within an identical visual token budget as the baseline.
Codes are open-sourced at \href{https://github.com/MAC-AutoML/QuoTA}{github link}.

\end{abstract}

\section{Introduction}
\label{sec:intro}


\begin{figure}[t!]
  \centering 
  \includegraphics[width=1.0\linewidth]{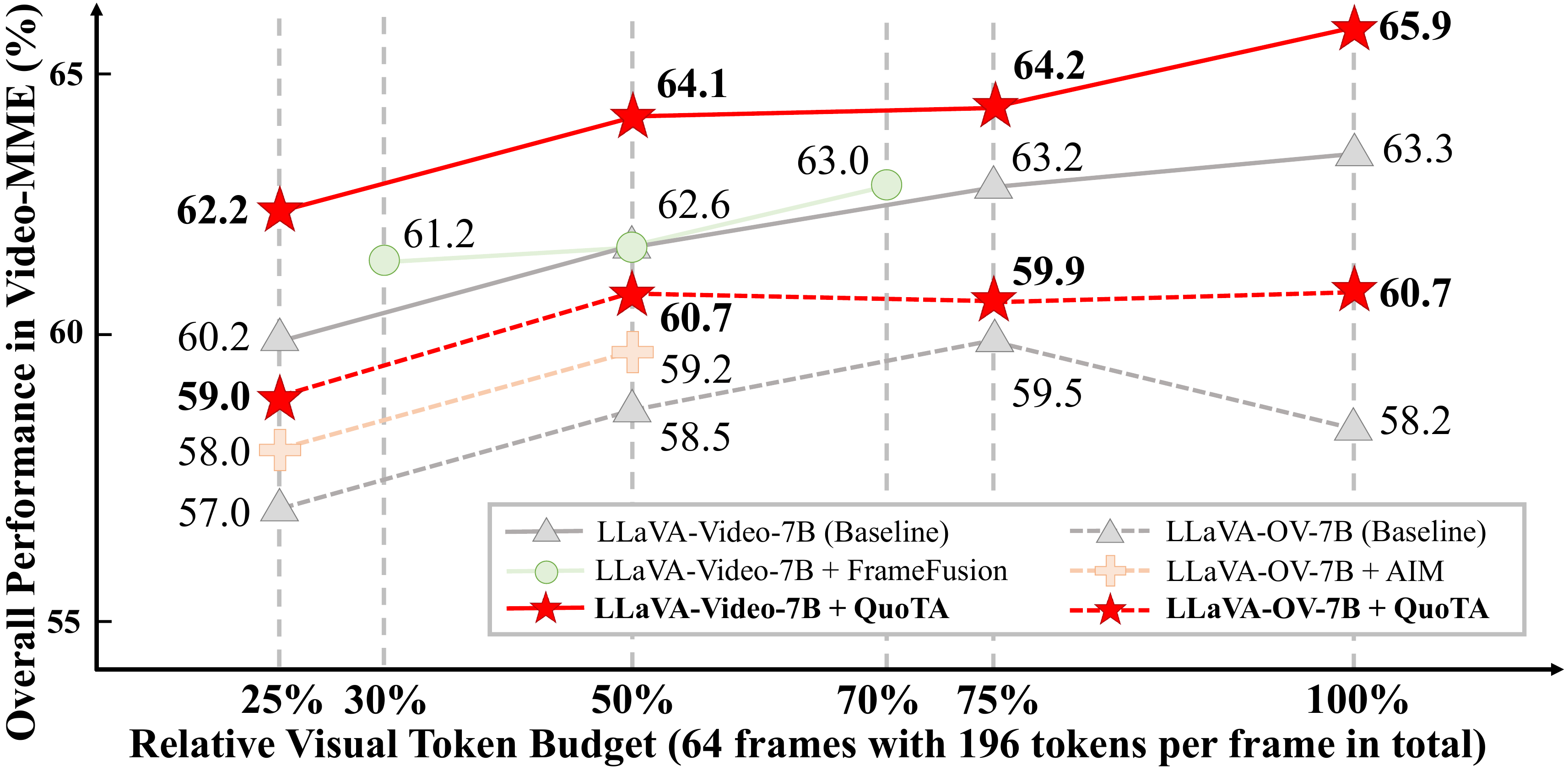}
  \caption{Comparative analysis of Video-MME \cite{videomme} when implementing attention-based token assignment methods AIM \cite{aim} and FrameFusion \cite{framefusion}, alongside our proposed query-oriented QuoTA within LLaVA-Video-7B \cite{llavavideo} and LLaVA-OV-7B \cite{llavaov} across varied relative visual token budgets. QuoTA demonstrates superior efficacy while exhibiting consistent performance enhancement across diverse token budget configurations relative to the baseline.}
  \label{fig_intro}
\end{figure}

With the emergence of advanced Large Language Models (LLMs), researchers have expanded their capabilities to video comprehension \cite{chatunivi, videochat, videollava, llavanextvideo, videollama, internvl, chen2024sharegpt4video}, establishing the domain of Large Video-Language Models (LVLMs).
Recent studies \cite{longva, longllava, intp, wang2024longllava, longvila} focus on extending LVLMs' reasoning context capacity, primarily through fine-tuning approaches for enhanced long video understanding.
However, empirical evidence from long-context LVLMs \cite{longva} and \cite{longvita} demonstrate performance degradation on Video-MME \cite{videomme} when frame sampling rates increase, suggesting that merely augmenting frame quantities introduces information redundancy while imposing greater computational demands on complex reasoning tasks.

Recent works \cite{aim, dycoke, framefusion, retake} have addressed visual token reduction through the analysis of attention value distributions across model layers, thereby providing more informationally efficient tokens for long video comprehension tasks. DyCoke \cite{dycoke}, for instance, demonstrates that attention score distributions among visual tokens exhibit significant sparsity, enabling the dynamic elimination of less-attended visual tokens within the KV cache.
However, as \textbf{post-hoc} methods that compress visual tokens either during or after their interaction with textual tokens within the decoder layers, they face substantial limitations:
\textbf{(i) Neglected task-specific token relevance.} 
Human visual cognition inherently focuses on query-relevant content when processing video-text tasks, highlighting the necessity of query-oriented token selection to align visual processing with task objectives while ensuring efficient token allocation and semantic coherence.
Conversely, visual tokens with high attention-weight responses primarily reflect inter-token associative strength rather than their direct pertinence to the query.  
As demonstrated by \cite{dyvte, aim}, cross-modal interactions mainly occur in early fusion stages, with later layers showing predominantly intra-modal patterns. Additionally, FrameFusion \cite{framefusion} reveals that token importance significance fluctuates across layers. These observations suggest that high-response visual tokens likely represent intra-modal relationships rather than query-relevant semantic features, potentially rendering attention-based token reduction counterproductive for task-specific comprehension tasks.
\textbf{(ii) Propagation of sequential reduction errors.} The hierarchical token reduction strategy based on attention patterns exhibits vulnerability to error accumulation, wherein suboptimal selection decisions implemented at initial layers adversely affect token selection in subsequent computational stages.

To address this issue, we propose QuoTA, an \textbf{ante-hoc} approach seamlessly integrated with existing LVLMs (termed \textit{based LVLM}) for visual tokens assignment before cross-modal interactions on decoder layers, enhancing query-specific visual token capture.
Specifically, QuoTA implements parallel video frame evaluation utilizing the zero-shot capabilities of an external lightweight LVLM (termed \textit{scoring LVLM}) to generate query-relevance scores, subsequently employing these metrics as discriminative criteria for visual token assignment.
Our query-oriented token assignment strategy enhanced cross-modal interactions while minimizing redundancy between visual and text tokens during early fusion phases, thereby augmenting performance.
To enhance scoring precision, we prompt the \textit{based LVLM} to decouple the query into a more interpretable question with Chain-of-Thoughts \cite{cot} reasoning. These reformulated questions subsequently prompt the \textit{scoring LVLM} in generating a query-specific relevance score for each sampled video frame.
After that, the visual tokens then proceed via one of three approaches based on the importance score: (i) bilinear interpolation, (ii) adaptive pooling, and (iii) dynamic token merging for spatial redundancy minimization.
Furthermore, since the elevated information density afforded by our query-oriented assignment protocol, we implement duration-dependent dynamic frame sampling (more sampling frames in longer videos and vice versa) to optimize the extraction of salient visual information. Notably, QuoTA can adapt to a given token budget.

We conduct experiments across diverse video understanding benchmarks, including Video-MME \cite{videomme}, MLVU \cite{mlvu}, LongVideoBench \cite{lvb}, VNBench \cite{vnbench}, MVBench \cite{mvbench}, and NeXT-QA \cite{nextqa}. 
Results demonstrate that LLaVA-Video-7B \cite{llavavideo} and LLaVA-OneVision-7B \cite{llavaov}, augmented with QuoTA in a plug-and-play manner, achieve an average performance improvement of 3.2\% and 2.5\% across all six benchmarks while maintaining equivalent computational requirements to their original baseline. 
Furthermore, as shown in Figure \ref{fig_intro}, QuoTA outperforms recent state-of-the-art approaches AIM \cite{aim} and FrameFusion \cite{framefusion} across varying visual token budgets when applied to LLaVA-Video-7B \cite{llavavideo} and LLaVA-OneVision-7B \cite{llavaov}, while maintaining consistent performance improvements over the baseline regardless of token budget, which is set to 64 frames with 196 tokens per frame (12,544 visual tokens in total).

In summary, our contributions are as follows:
\begin{itemize}
    \item \textbf{We design a versatile plug-and-play pipeline for existing LVLMs:} QuoTA provides a training-free solution applicable to diverse LVLMs, enhancing long video understanding performance by assigning visual tokens based on text instruction (query) relevance. This approach offers a more elegant and direct methodology compared to conventional attention-based analytical techniques.
    \item \textbf{We propose CoT-driven query decouple for query-oriented frame scoring:} QuoTA employs Chain-of-Thoughts to decouple query into a specific-designed question, enabling high-quality scoring of video frames.
    \item \textbf{Our QuoTA setting a new state-of-the-art:} Integration of QuoTA with LLaVA-Video-7B yields a 3.2\% average performance improvement across six benchmarks, achieving the best results in five video benchmarks, including Video-MME and MLVU, among 7B LVLMs.
\end{itemize}

\section{Related Work}
\label{sec:related_work}

\begin{figure*}[t!]
  \centering 
  \includegraphics[width=1.0\linewidth]{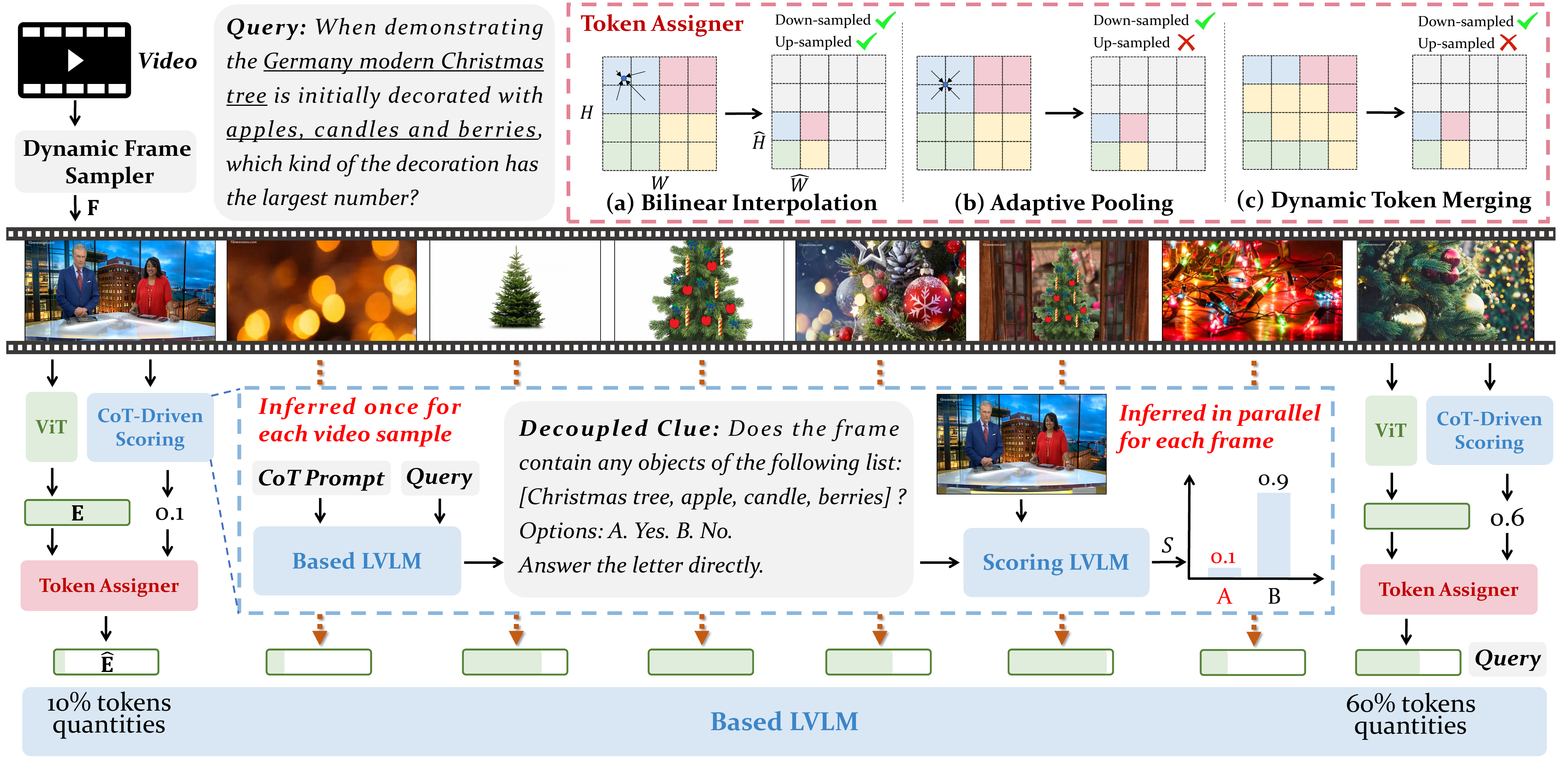}
  \caption{The framework of QuoTA. Initially, a dynamic frame sampler extracts $T$ frames from the video based on its duration, which are subsequently processed by \texttt{ViT} to generate visual embeddings $\bm{\mathrm{E}}$. Then, the based LVLM decouples the input query using Chain-of-Thoughts \cite{cot} reasoning into a decoupled clue to generate frame-wise importance scores through scoring LVLM in parallel, thus evaluating the relevance to the query of each frame. Finally, a token assigner rescales the frame embeddings to $\bm{\mathrm{\hat{E}}}$ based on these importance scores.}
  \label{fig_framework} 
\end{figure*}

\subsection{Large Video-Language Models}

Recent advances in large language models (LLMs) have sparked interest in developing video understanding systems. Video-ChatGPT \cite{videochatgpt} processes videos by extracting frame-level features through spatial and temporal pooling. 
VideoChat \cite{videochat} combines textual descriptions with video appearance embeddings.
Video-LLaVA \cite{videollava} aligns image and video encoders using a shared projector to map representations into a common language space.
LLaVA-NeXT-Video \cite{llavanextvideo} extends LLaVA-NeXT \cite{llavanext} through video-specific fine-tuning.
While these models show promise, they still struggle with analyzing long-form videos.

\subsection{Long Video Understanding}

Recent research has focused on expanding context window sizes for long video understanding.
LongVA \cite{longva}, Video-XL \cite{videoxl}, and LongVILA \cite{longvila} leverage LLMs' long-text comprehension capabilities through continuous training. However, they may suffer from performance degradation with excessive frame sampling due to video content redundancy and model capacity constraints.
Several works \cite{wang2024videoagent, gupta2023visual, suris2023vipergpt, video-rag} have employed LLMs with tools to process video as structured text for question-answering.
However, these methods are limited by extensive processing times and reliance on proprietary models (e.g., GPT-4o).

\subsection{Efficient Visual Modeling}

Previous research in multi-modal LLMs assigns visual tokens to reduce spatial redundancy. FastV \cite{fastv} reduces them at a particular selected layer based on the distribution of attention.
Recent works \cite{framefusion, dycoke, aim, retake} extend it by reducing the redundancy of spatiotemporal information for long video understanding. For example, AIM \cite{aim} merges similar tokens and then preserves the important tokens hierarchically.
However, they rely on attention weight analysis, which only reflects token-wise associations rather than query-specific relevance; this indirect measurement may not accurately capture task-specific token importance. 
Moreover, as mentioned in FrameFusion \cite{framefusion}, token importance is inconsistent across different layers; thus, hierarchical reduction is susceptible to error propagation from early suboptimal decisions.
Other works \cite{longvu, llavamini, dynamicvlm, PVC, videochatflash, rtl, vtm} integrate token assignment into the architecture design and enhance model capabilities through training. 
However, the training-needed schema makes them inflexible and less generalizable.

\section{Method}
\label{sec:formatting}

\subsection{Preliminaries}

For a given video $\bm{\mathrm{V}}$, a frame sampler extracts $T$ frames $\bm{\mathrm{F}} = \{\bm{\mathrm{F}}_i\}_{i=1}^T$. Conventional approaches typically employ uniform sampling at fixed intervals for computational efficiency. Features of each frame are then obtained via $\bm{\mathrm{E}} = \{\bm{\mathrm{E}}_i\}_{i=1}^T = \texttt{ViT}(\bm{\mathrm{F}})$, where $\texttt{ViT}$ represents a transformer-based visual encoder (e.g., CLIP-L \cite{clip}), and $\bm{\mathrm{E}}_i\in\mathbb{R}^{(H \times W) \times C}$ denotes the visual tokens corresponding to the $i$-th frame, with $H$, $W$ representing spatial dimensions and $C$ the feature dimension.
In standard LVLM, the video features $\bm{\mathrm{E}}$ alongside the user query $\bm{\mathrm{Q}}$ are processed to generate the output $\bm{\mathrm{O}} = \texttt{LVLM}(\bm{\mathrm{E}}, \bm{\mathrm{Q}})$. Our proposed QuoTA enhances this process by first assessing frame-level importance, which subsequently guides token assignment within each frame to produce refined features $\bm{\mathrm{\hat{E}}}$. The final output is thus formulated as $\bm{\mathrm{O}} = \texttt{LVLM}(\bm{\mathrm{\hat{E}}}, \bm{\mathrm{Q}})$. Figure \ref{fig_framework} illustrates the overall architecture of our approach.

\subsection{LVLM-Based Frame Scoring}
\label{subsec:3_2}

Given the inherent information redundancy in sampled video frames, it is advantageous to compress query-irrelevant visual tokens while directing the LVLM's attention toward keyframes. 
While text-image similarity scores from CLIP \cite{clip} could theoretically assess query-frame relevance, CLIP's known bias toward physical entity nouns often yields suboptimal frame importance assessments in reasoning scenarios with a complex query.
Instead, in QuoTA, we leverage the robust multi-choice reasoning capabilities of LVLMs by formulating a binary choice question for each frame. This question is processed by a lightweight scoring LVLM, which produces an importance score $S$ (a scalar) derived from the probability of selecting option ``A". The binary-choice prompt structure fed into the scoring LVLM for each frame follows the format:

\begin{table}[h]
\centering
\begin{tabular}{|p{0.9\linewidth}|}
\hline
\rowcolor[gray]{0.8}
\textbf{Prompt for Frame Scoring Based on Source Query} \\ \hline
\rowcolor[gray]{0.9}
Question: Does this frame contain any information to answer the given query: \{query\}? \\
\rowcolor[gray]{0.9}
\noindent A. Yes. B. No. \\
\rowcolor[gray]{0.9}
Answer the letter directly. \\
\hline
\end{tabular}
\end{table}

To optimize the trade-off between scoring accuracy and computational efficiency, we employ Qwen2-VL-2B \cite{qwen2vl} as our lightweight scoring LVLM, which facilitates \textbf{parallel inference} across all video frames with minimal latency and modest GPU resource requirements. Upon obtaining the importance score for each sampled frame, we normalize these values to serve as allocation guides for token assignment. The normalized importance score for each frame is computed as $S_n = \{S_n^i\}_{i=1}^T = S^i / \sum_{j=1}^T S^j$.

\subsection{CoT-Driven Query Decouple}
\label{subsec:3_3}

Employing the prompt described in Section \ref{subsec:3_2} yields only marginal improvements, as the scoring LVLM tends to optimistically presume that frames inherently contain adequate information. This results in homogeneous importance weights across frames, compromising discriminative capacity. To address this limitation, we implement a Chain-of-Thoughts \cite{cot} approach that decouples the query that enhances frame differentiation quality, as validated in our ablation studies (Section \ref{subsec:4_4}).
Considering that LVLMs demonstrate superior capability in identifying physical entities within visual content, QuoTA utilizes the based LVLM to decouple the original query into a structured object list. 
Specifically, we implement a three-step prompting protocol (the prompt is detailed in the Appendix) that directs the base LVLM to decouple the query, extracting concrete physical entities to interrogate the scoring LVLM, which markedly reduces hallucination. This Chain-of-Thoughts process encompasses (i) assessing the necessity for entity decoupling, (ii) transforming the original query into a structured object list when appropriate, and (iii) refining this list by eliminating abstract concepts. When the based LVLM determines entity recognition is warranted, we query the scoring LVLM using the following binary-choice prompt:

\begin{table}[h]
\centering
\begin{tabular}{|p{0.9\linewidth}|}
\hline
\rowcolor[gray]{0.8}
\textbf{Prompt for Frame Scoring Based on Entity List} \\ \hline
\rowcolor[gray]{0.9}
Question: Does the frame contain any objects of the following list: \{object\_list\}? \\
\rowcolor[gray]{0.9}
\noindent A. Yes. B. No. \\
\rowcolor[gray]{0.9}
Answer the letter directly. \\
\hline
\end{tabular}
\end{table}

\noindent Otherwise, we leverage the prompt in Section \ref{subsec:3_2}. We also conduct other CoT-driven decouple strategies for query-oriented keyframes scoring, which details in Section \ref{subsec:4_4}.

\subsection{Dynamic Visual Token Assignment}
\label{subsec:3_4}

After obtaining the normalized scores $S_n = \{S_n^i\}_{i=1}^T$ for all frames, we calculate the target token quantities $N = \{N_i\}_{i=1}^T$ for each frame, where $N_i = S_n^i \times N_t$, with $N_t$ representing the total visual token budget. This budget can be user-defined to accommodate specific computational resource constraints. Notably, individual frame token quantities $N_i$ may exceed $N_t / T$.
We established $N_t$ according to the empirically optimal frame configuration of the based LVLM to ensure experimental consistency. For instance, LLaVA-Video \cite{llavavideo} demonstrates optimal performance with 64 frames at 196 tokens per frame; consequently, $N_t$ was standardized at $64 \times 196 = 12,544$. This methodological decision was implemented for multiple reasons: (i) Equivalent token counts enable direct comparative assessment of QuoTA's efficacy while controlling computational variables; (ii) Since the base LVLM was trained with this optimal frame configuration that generalizes effectively across most scenarios, maintaining consistent total token budget typically enhances performance and transferability.

Subsequently, we assign visual tokens for each frame according to the target token quantities $N$ using a dynamic token assigner. In this study, we examine three distinct categories of dynamic token assigners as follows:

\textbf{(1) Bilinear Interpolation.} A straightforward approach involves employing bilinear interpolation to resize the feature maps. For the $i$-th input frame embeddings $\bm{\mathrm{E}}_i\in \mathbb{R}^{(H \times W) \times C}$ and its corresponding target token quantities $N_i$, we compute optimal spatial dimensions $\hat{H}_i$ and $\hat{W}_i$ that satisfy $\hat{H}_i \times \hat{W}_i$ being closest to but not exceeding $N_i$:
\begin{equation}
\begin{aligned}
\hat{H}_i = \hat{W}_i &= \lfloor\sqrt{N_i}\rfloor \\
\text{if } \hat{H}_i \times \hat{W}_i &< N_i: \\[-3pt]
&\!\!\!\!\begin{cases}
\hat{H}_i \leftarrow \hat{H}_i + 1 & \text{if } (\hat{H}_i + 1) \times \hat{W}_i \leq N_i \\
\hat{W}_i \leftarrow \hat{W}_i + 1 & \text{otherwise.} 
\end{cases}
\end{aligned}
\label{equa:1}
\end{equation}
Then, bilinear interpolation is applied to transform the original embeddings $\bm{\mathrm{E}}_i$ into $\hat{\bm{\mathrm{E}}_i} \in \mathbb{R}^{(\hat{H}_i\times \hat{W}_i) \times C}$:
\begin{equation}
\hat{\bm{\mathrm{E}}_i} = \texttt{B\_Interpolate}(\bm{\mathrm{E}}_i, [\hat{H}_i, \hat{W}_i])
\end{equation}

\textbf{(2) Adaptive Pooling. } Application of pooling operations represents an intuitive approach. However, as a down-sampling technique, it requires the target token quantities to satisfy $N_i \le H \times W$. Consequently, the normalized weights $S_n$ require further processing. For the $i$-th input frame whose target token allocation exceeds spatial constraints (i.e., $N_i > H \times W$), we redistribute their excess weights to other frames proportionally. Let $\mathcal{L} = \{i | S_n^i \cdot N > H \times W\}$ denote frames exceeding the limit, the adjusted weights are:
\begin{equation}
\hat{S_n^i} = \begin{cases}
\frac{H \times W}{N_t} & \text{if } i \in \mathcal{L} \\
S_n^i + \frac{S_n^i}{\sum_{j \notin \mathcal{L}} S_n^j} \sum_{k \in \mathcal{L}} (S_n^k - \frac{H \times W}{N_t}) & \text{otherwise}
\end{cases}
\end{equation}

The adjusted target quantities of tokens for the $i$-th frame is subsequently derived as $\hat{N_i} = \hat{S_n^i} \times N_t$. We compute optimal spatial dimensions $\hat{H}_i$ and $\hat{W}_i$, employing the methodology delineated in Equation \ref{equa:1}.
Finally, adaptive pooling is applied to transform the original embeddings $\bm{\mathrm{E}}_i$ into $\hat{\bm{\mathrm{E}}_i}$:
\begin{equation}
\hat{\bm{\mathrm{E}}_i} = \texttt{A\_AvgPool2d}(\bm{\mathrm{E}}_i, [\hat{H}_i, \hat{W}_i])
\end{equation}

\textbf{(3) Dynamic Token Merging.} It implements a token merging (down-sampling) operation predicated on cosine similarity metrics between distinct tokens within a visual representation, as introduced in ToMe \cite{tome}. We compute  the optimal spatial dimensions $\hat{H}_i$ and $\hat{W}_i$ utilizing methodologies analogous to Adaptive Pooling, subsequently apply the $k$-th Bipartite Soft Matching alternately along rows and columns to transform the original embeddings $\bm{\mathrm{E}}_i$ into $\hat{\bm{\mathrm{E}}_i}$:
\begin{equation}
\hat{\bm{\mathrm{E}}_i} = \texttt{B\_SoftMatching}(\bm{\mathrm{E}}_i, [\hat{H}_i, \hat{W}_i])
\end{equation}

\subsection{Dynamic Frame Sampling}
\label{subsec:3_5}

In QuoTA, considering that the information redundancy in the video can be mitigated through dynamic visual token assignment, we accommodate additional input frames within the LVLM to capture more potential critical content.
Consequently, we implement uniform frame sampling with adaptive quantity parameters determined by video duration. Specifically, for a video spanning $t$ seconds, the sampled frame count $T$ is calculated according to:
\begin{equation}
T = T_{base} + \min(\lfloor \frac{t}{3600} \times \alpha \rfloor, \alpha)
\end{equation}
where $T_{base}$ is the base number of frames (e.g., 96), and $\alpha$ is a hyperparameter that controls the upper bound of additional frames (e.g., 64). This formulation ensures that long video sequences receive proportionally increased sampling density to capture salient information while maintaining computational efficiency by capping the maximum additional frames at $\alpha$. The frames are then uniformly sampled across the video timeline at intervals of $t/T$ seconds.
\section{Experiments}
\label{sec:exp}

\begin{table*}[t!]
\centering
\setlength{\tabcolsep}{1.8mm}
\renewcommand\arraystretch{1.0} 
\begin{tabular}{l|cc|c|c|cccc}
\toprule
\multirow{2}{*}{\textbf{Model}}   & \multirow{2}{*}{\textbf{Params}} & \multirow{2}{*}{\textbf{Frames}} & \textbf{LongVideo} & \textbf{MLVU} & \multicolumn{4}{c}{\textbf{Video-MME} (wo/w-subtitles)} \\ 
 &  &  & \textbf{Bench} (val) & (m-avg) & Short & Medium & Long & Overall  \\
  \midrule
    \multicolumn{9}{c}{\textbf{\textit{Proprietary LVLMs}}} \\
\midrule
\rowcolor{gray!20} GPT-4o \cite{gpt4o} & - & 384 & 66.7 & 64.0 & 80.0/82.8 & 70.3/76.6 & 65.3/72.1 & 71.9/77.2 \\
\rowcolor{gray!20} Gemini-1.5-Pro \cite{gemini} & - & 0.5 fps & 64.0 & - & 81.7/84.5 & 74.3/81.0 & 67.4/77.4 & 75.0/81.3 \\
  \midrule
      \multicolumn{9}{c}{\textbf{\textit{Open-Source LVLMs}}} \\
      \midrule
LongVA \cite{longva} & 7B & 128 & - &  56.3 & 61.1/61.6 & 50.4/53.6 & 46.2/47.6 & 52.6/54.3 \\
Video-XL \cite{videoxl} & 7B & 128 & 50.7 & 64.9 & 64.0/67.4 & 53.2/60.7 & 49.2/54.9 & 55.5/61.0 \\
VITA-1.5 \cite{fu2025vita} & 7B & 16 & - & - & 67.0/69.9 & 54.2/55.7 & 47.1/50.4 & 56.1/58.7 \\
TimeMarker \cite{timemarker} & 8B & 128 & 56.3 & 63.9 & 71.0/75.8 & 54.4/60.7 & 46.4/51.9 & 57.3/62.8 \\
AIM \cite{aim} (LLaVA-OV) & 7B & 32* & - & 69.3 & -/- & -/- & -/- & 59.2/62.3 \\
LongVILA \cite{longvila} & 7B & 256 & 57.1 & - & 69.0/72.9 & 58.3/64.9 & 53.0/57.4 & 60.1/65.1 \\
LongVU \cite{longvu} & 7B & - & - & 65.4 & -/- & -/- & 59.5/- & 60.6/- \\ 
Qwen2-VL \cite{qwen2vl} & 7B & - & - & 64.8 & -/- & -/- & -/- & 63.3/69.0 \\
ReTaKe \cite{retake} (Qwen2-VL) & 7B & - & - & 69.8 & 72.8/- & 62.7/- & 56.2/- & 63.9/- \\
NVILA \cite{nvila} & 8B & 256 & 57.7 & 70.1 & \underline{75.7}/\underline{77.6} & 62.2/\textbf{69.0} & \underline{54.9}/63.3 & \underline{64.2}/\textbf{70.0} \\
\midrule
LLaVA-OV \cite{llavaov}  & 7B & 64 & 56.3 & 64.7 & 70.2/74.0 & 56.6/64.2 & 47.7/62.4 & 58.1/66.9 \\
\rowcolor{cyan!10} LLaVA-OV + QuoTA  & 7B & 64* & 57.4 & 69.7 & 71.1/74.8 & 58.8/65.2 & 52.2/\textbf{63.9} & 60.7/68.0 \\
\midrule
LLaVA-Video \cite{llavavideo} & 7B & 64 & \underline{58.2} & \underline{70.8} & 75.4/77.3 & \underline{62.6}/67.7 & 51.8/\underline{63.6} & 63.3/\underline{69.5} \\
\rowcolor{cyan!10} LLaVA-Video + QuoTA  & 7B & 64* & \textbf{59.0} & \textbf{71.9} & \textbf{77.1}/\textbf{79.0} & \textbf{64.9}/\underline{68.0} & \textbf{55.7}/62.9 & \textbf{65.9}/\textbf{70.0}  \\  
  \bottomrule
\end{tabular}
\caption{Performance on the validation set of LongVideoBench \cite{lvb}, MLVU \cite{mlvu} and Video-MME \cite{videomme}. By applying QuoTA to LLaVA-Video-7B \cite{llavavideo}, we observed an average performance improvement of 1.5\% across three long video understanding benchmarks while setting new state-of-the-art. * denotes using the same visual token budget as the baseline. Models in parentheses represent the baselines they used.}
\label{tab_main}
\end{table*}

\subsection{Datasets}

To ensure robustness, we evaluated QuoTA across six datasets: \textbf{Video-MME} \cite{videomme}: A widely used benchmark for assessing the ability of LVLMs to handle detailed videos in real-world scenarios that varying lengths (short, medium, long). \textbf{MLVU} \cite{mlvu}: A large-scale long video benchmark with a large wide of 9 distinct tasks and diversified lengths, ranging from 3 minutes to 2 hours. \textbf{LongVideoBench} \cite{lvb}: A benchmark designed to accurately retrieve and reason over detailed multimodal information from long videos with 17 fine-grained categories. \textbf{VNBench} \cite{vnbench}: A synthetic benchmark designed to evaluate models’ long-context abilities, covering tasks such as retrieval, ordering, and counting. \textbf{MVBench} \cite{mvbench}: A benchmark cross over 20 challenging video understanding tasks, focusing on temporal understanding in dynamic video tasks. \textbf{NeXT-QA} \cite{nextqa}: A short-video benchmark emphasizing causal and temporal reasoning, challenging models to understand complex sequences.

\subsection{Implementation Details}

We performed all the experiments on NVIDIA A100 40G GPUs.
We extend LLaVA-Video \cite{llavavideo} and LLaVA-OneVision \cite{llavaov} with our QuoTA at 7B-scale, constrained by available computational resources.
Bilinear Interpolation is our dynamic token assigner, offering enhanced flexibility in up- and down-sampling operations while demonstrating superior performance, as evidenced in Table \ref{abl_token}.
For equitable comparative analysis, we align the total visual tokens budget $N_t$ with each LVLM's original total visual token quantities during inference.
To optimize the quality-efficiency trade-off, we use Qwen2-VL-2B \cite{qwen2vl} as our scoring LVLM. The base frame quantity $T_{base}$ is configured at 96 while the maximum additional frame $\alpha$ is set to 64 across all benchmarks except for VNBench \cite{vnbench} ($T_{base}=128$, $\alpha=96$).

\subsection{Main Results}

We evaluate QuoTA implemented within LLaVA-Video \cite{llavavideo} and LLaVA-OneVision \cite{llavaov} at 7B-scale, maintaining equivalent computational constraints (total visual tokens budget $N_t$) as the baseline across three long video understanding benchmarks: LongVideoBench \cite{lvb}, MLVU \cite{mlvu} and Video-MME \cite{videomme}.
The empirical outcomes presented in Table \ref{tab_main} demonstrate that QuoTA integration into LLaVA-Video-7B \cite{llavavideo}, yielding improvements of 0.8\%, 1.1\%, and 2.6\% on LongVideoBench \cite{lvb}, MLVU \cite{mlvu}, and Video-MME \cite{videomme} (w/o subtitles), respectively.
Notably, substantial performance enhancements manifest in extended-duration video (spanning 30-60 minutes) within Video-MME \cite{videomme} (47.7\% → 52.2\% for LLaVA-OneVision \cite{llavaov}, and 51.8\% → 55.7\% for LLaVA-Video \cite{llavavideo}) under ``w/o subtitles'' conditions, substantiating that our query-oriented token assignment methodology together with dynamic frame sampling strategy effectively mitigates information redundancy (particularly pronounced in long videos) while accentuating salient content, thereby facilitating enhanced model activation and comprehension of complex visual narratives.

Furthermore, we conduct evaluations of QuoTA on two conventional video understanding benchmarks, MVBench \cite{mvbench} and NeXT-QA \cite{nextqa}, alongside the specifically constructed Needle-In-A-Haystack video benchmark VBNench \cite{vnbench}, as illustrated in Table \ref{tab_sub}. 
Particularly noteworthy is the substantial enhancement observed on VNBench \cite{vnbench} (44.7\% → 49.3\% for LLaVA-OneVision \cite{llavaov} and 54.4\% → 64.7\% for LLaVA-Video \cite{llavavideo}), empirically validating that our query-oriented frame-wise scoring methodology effectively directs the LVLM's attention toward query-relevant keyframes. 
Additionally, as shown by Figure \ref{fig_intro}, our proposed QuoTA demonstrates superior efficacy that outperforms recent SoTAs, FrameFusion \cite{framefusion} and AIM \cite{aim} at distinct relative visual token budget allocations.
Notably, QuoTA establishes new SoTA across five benchmarks.

\begin{table}[]
\centering
\setlength{\tabcolsep}{0.5mm}
\renewcommand\arraystretch{0.95}
\begin{tabular}{l|ccc}
\toprule
\textbf{Model} & \textbf{MVBench} & \textbf{NeXT-QA} & \textbf{VNBench} \\ \midrule
\multicolumn{4}{c}{\textbf{\textit{Proprietary LVLMs}}} \\ \midrule
\rowcolor{gray!20} Gemini-1.5-Pro \cite{gemini} & - & - & 66.7 \\
\rowcolor{gray!20} GPT-4o \cite{gpt4o} & - & - & 64.4 \\
\midrule
\multicolumn{4}{c}{\textbf{\textit{Open-Source LVLMs}}} \\ \midrule
LongVA \cite{longva} & - & 68.3 & 41.5 \\
mPLUG-Owl3 \cite{mplugowl3} & 54.5 & 76.8 & - \\
Video-XL \cite{videoxl} & 55.3 & - & 61.6 \\
LongVILA \cite{longvila} & \textbf{67.1} & 80.7 & \underline{63.0} \\
\midrule
LLaVA-OV \cite{llavaov} & 56.7 & 79.4 & 44.7 \\
\rowcolor{cyan!10} LLaVA-OV + QuoTA & 57.3 & 80.4 & 49.3 \\ \midrule
LLaVA-Video \cite{llavavideo} & 58.6 & \underline{83.2} & 54.4 \\
\rowcolor{cyan!10} LLaVA-Video + QuoTA & \underline{62.1} & \textbf{83.9} & \textbf{64.7} \\

\bottomrule
\end{tabular}
\caption{The overall performance on MVBench \cite{mvbench}, VNBench \cite{vnbench} and NeXT-QA \cite{nextqa} at 7B-scale LVLMs with the setting of the original frame rates. By applying QuoTA to LLaVA-Video-7B \cite{llavavideo}, we observed an average performance improvement of 4.8\% across three benchmarks, especially a 10.3\% improvement on the Needle-In-A-Haystack benchmark VNBench \cite{vnbench}, which set a new state-of-the-art performing better then LongVILA \cite{longvila}, demonstrating QuoTA assist query-oriented keyframes focusing.}
\label{tab_sub}
\end{table}

\subsection{Ablation Studies}
\label{subsec:4_4}

\textbf{Effect of different components of QuoTA.} 
To investigate the efficacy of QuoTA's components, we conduct ablation experiments with varied configurations to evaluate LLaVA-Video-7B \cite{llavavideo} performance on Video-MME \cite{videomme} and VNBench \cite{vnbench}.
As shown in Table \ref{abl_comp}, when compared to fixed-length sampling (96-frames with $\sim$131 visual tokens per frame, maintaining equivalent token budget $N_t$ as the baseline of 64-frames with 196 visual tokens per frame), dynamic-length sampling (Section \ref{subsec:3_5}, employing adaptive frame sampling 96$\sim$160 frames while preserving consistent token budget $N_t$) exhibits superior performance only when augmented with our proposed LVLM-based frame scoring methodology detailed in Section \ref{subsec:3_2} for visual token assignment.
Additionally, implementing the Chain-of-Thoughts-driven query decoupling technique outlined in Section \ref{subsec:3_3} yields substantial performance enhancement by improving scoring precision, especially in VNBench \cite{vnbench}.
These empirical findings indicate that increasing frame sampling without discriminative selection provides limited improvement due to query-irrelevant information redundancy.

\begin{table}[]
\centering 
\setlength{\tabcolsep}{1.0mm} 
\renewcommand\arraystretch{0.95} 
\begin{tabular}{cc|cc|cc}
\toprule
 \textbf{Fix-len.} & \textbf{Dy-len.} & \textbf{Wei.} & \textbf{CoT-Dec.} & \textbf{V-MME} & \textbf{VNBench}   \\ \midrule
 &  &   &  & 63.3 & 54.4 \\ 
\checkmark &  &   &  & 64.0 & 58.7 \\
 & \checkmark &   &  & 63.5 & 58.4 \\
\checkmark &  &  \checkmark &  & 63.6 & 49.0 \\
 & \checkmark &  \checkmark &  & \underline{64.4} & 48.6 \\
\checkmark &  &  \checkmark & \checkmark & 64.2 & \textbf{60.9} \\
\rowcolor{cyan!10}  & \checkmark &  \checkmark & \checkmark & \textbf{65.9} & \underline{60.6} \\
  \bottomrule
\end{tabular}
\caption{Results on combinations of different components in Video-MME \cite{videomme} and VNBench \cite{vnbench} when using Long-LLaVA-7B \cite{longllava} as the based LVLM on QuoTA. \textbf{Fix-len.} and \textbf{Dy-len.} represent fix sampled 96-frame and dynamic sampled 96$\sim$160 frames with the same token budget $N_t$ as the baseline, respectively. \textbf{Wei.} and \textbf{CoT-Dec.} denote the LVLM-based frame scoring and CoT-Driven Query Decouple in Section \ref{subsec:3_3} and \ref{subsec:3_4}, respectively. }
\label{abl_comp}
\end{table}

\noindent \textbf{Effect of different frame-sampling strategies and visual token budget.} 
As mentioned in Section \ref{subsec:3_4} and \ref{subsec:3_5}, we include serval hyper-parameters in QuoTA: (i) The budget of total visual tokens $N_t$, (ii) the base number $T_{base}$, and (iii) the maximum additional quantities $\alpha$. 
We conduct experiments with different settings of these hyper-parameters when extending LLaVA-Video-7B \cite{llavavideo} with QuoTA on Video-MME \cite{videomme} and VNBench \cite{vnbench}. 
We carefully examined the trade-offs between accuracy and inference time: while larger values of $T_{base}$ and $\alpha$ generally lead to better performance, they also increase the inference latency proportionally. 
As shown in Table \ref{abl_frame}, setting $T_{base}=96$ and $\alpha=64$ offers an optimal balance, providing substantial accuracy gains without excessive computational overhead in both Video-MME \cite{videomme} and VNBecnh \cite{vnbench}. While QuoTA achieves state-of-the-art performance (64.7\%) on VNBench \cite{vnbench} with $T_{base}=128$ and $\alpha=96$, it incurs significantly higher computational overhead.
At 50\% token budget ($T_{base}= 96$, $\alpha=64$), QuoTA achieves 64.1\% accuracy, surpassing the full-token baseline (63.3\%) while reducing inference time. With a more aggressive reduction to 25\% token budget ($T_{base}= 64$, $\alpha=32$), it maintains competitive performance with only a 1.1\% accuracy drop while achieving 2.1$\times$ inference speedup.

\begin{table}[]
\centering 
\setlength{\tabcolsep}{1.5mm} 
\renewcommand\arraystretch{0.95} 
\begin{tabular}{cccc|cc}
\toprule
 \textbf{$T_{base}$} & \textbf{$\alpha$} & \textbf{$N_t$} & \textbf{\#Time} & \textbf{V-MME} & \textbf{VNBench}  \\ \midrule
\rowcolor{gray!10} 16 & 0 & 3,136 & 4.50s & 60.0 & 30.7 \\ 
\rowcolor{gray!10} 32 & 0 & 6,272 & 9.07s & 62.6 & 40.5 \\
\rowcolor{gray!10} 64 & 0 & 12,544 & 15.31s & 63.3 & 54.4 \\ \midrule
\multicolumn{6}{c}{\textit{Dynamic sampling with 25\% budget (3,136 tokens)}} \\ \midrule
 32 & 32 & 3,136 & 6.59s & 61.4 & 39.0 \\
  \rowcolor{cyan!10} 64 & 32 & 3,136 & 7.42s & \textbf{62.2} & 49.2 \\
 64 & 64 & 3,136 & 8.33s & 61.5 & \underline{49.9} \\
 96 & 64 & 3,136 & 9.54s & \underline{62.0} & \textbf{50.1} \\ \midrule
\multicolumn{6}{c}{\textit{Dynamic sampling with 50\% budget (6,272 tokens)}} \\ \midrule
 32 & 32 & 6,272 & 11.51s & 62.3 & 35.0 \\
 64 & 32 & 6,272 & 12.34s & 63.3 & 47.5 \\
 64 & 64 & 6,272 & 13.29s & 63.8 & 54.6 \\
 \rowcolor{cyan!10} 96 & 64 & 6,272 & 14.13s & \textbf{64.1} & \textbf{58.0} \\
 96 & 96 & 6,272 & 15.07s & \underline{64.0} & \underline{57.9} \\ \midrule
\multicolumn{6}{c}{\textit{Dynamic sampling with 100\% budget (12,544 tokens)}} \\ \midrule
 64 & 64 & 12,544 & 20.85s & 64.6 & 56.0 \\
 \rowcolor{cyan!10} 96 & 64 & 12,544 & 21.68s & \textbf{65.9} & 60.6 \\
 96 & 96 & 12,544 & 22.97s & 64.6 & 60.9 \\
 128 & 96 & 12,544 & 24.82s & 64.1 & \textbf{64.7} \\
 128 & 128 & 12,544 & 26.49s & \underline{64.7} & \underline{63.7} \\
  \bottomrule
\end{tabular}
\caption{Results on different frame-sampling strategies and token budgets in Video-MME \cite{videomme} and VNBench \cite{vnbench}. Gray rows represent baseline LLaVA-Video-7B \cite{llavavideo} with fixed frame sampling while others represent dynamic sampling with varying $T_{base}$ and $\alpha$ in different budget $N_t$ when extending the baseline with QuoTA. \textbf{\#Time} denotes the average time during inference per sample.}
\label{abl_frame}
\end{table}

\begin{figure*}[h!]
  \centering 
  \includegraphics[width=1.0\linewidth]{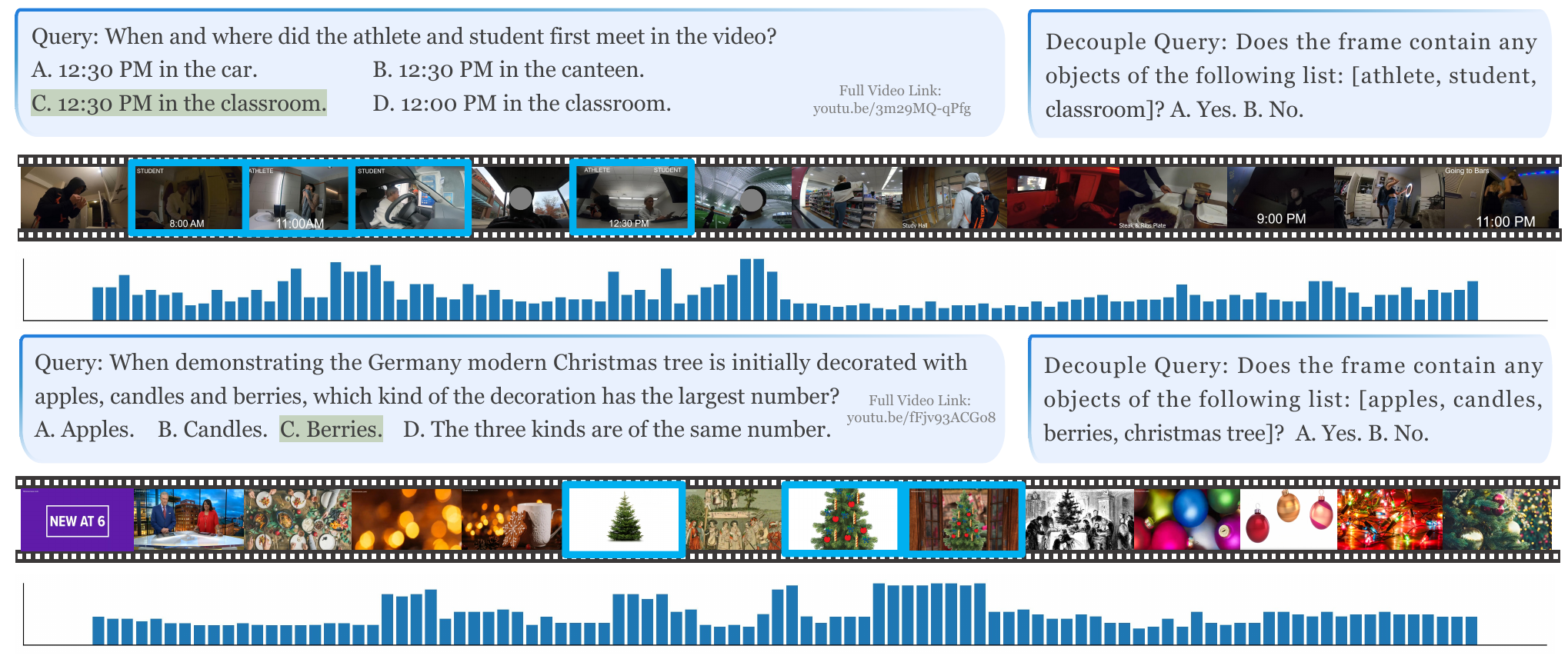}
  \caption{Qualitative result shown in Video-MME \cite{videomme} benchmark when applying QuoTA with LLaVA-Video-7B \cite{llavavideo}. The video frames with a blue border are query-oriented keyframes, and the bar chart shows the normalized scores of QuoTA for each frame.}
  \label{fig_exam} 
\end{figure*}

\noindent \textbf{Effect of different visual token assignment strategy.} 
In Section \ref{subsec:3_4}, we proposed three types of dynamic visual token assigners. To evaluate their comparative efficacy, we conducted an ablation study implementing QuoTA within LLaVA-Video-7B \cite{llavavideo} on the Video-MME \cite{videomme} and VNBench \cite{vnbench}. As detailed in Table \ref{abl_token}, optimal performance is achieved with bilinear interpolation. Despite the token similarity-based merging approach employed by ToMe \cite{tome}, we contend that such a methodology disrupts spatial coherence in video representations, consequently impeding effective cross-modal interaction during early fusion stages.
Similarly, adaptive pooling encounters analogous limitations, as it compromises continuous spatial structure within visual features, potentially degrading spatial attention quality and cross-modal alignment precision. 
Conversely, bilinear interpolation provides flexibility, supporting both up and down-sampling operations while preserving spatial continuity and maintaining stable cross-modal information propagation, facilitating robust feature learning and cross-modal integration. 
The results further suggest that prevalent attention-based token assignment methodologies, which prioritize token similarity for merging operations, may represent suboptimal strategic approaches.

\begin{table}[]
\centering 
\setlength{\tabcolsep}{0.7mm} 
\renewcommand\arraystretch{0.95} 
\begin{tabular}{l|cccc|c}
\toprule
\multirow{2}{*}{\textbf{Visual Token Assigner}} & \multicolumn{4}{c|}{\textbf{Video-MME}} & \multirow{2}{*}{\textbf{VNBench}}   \\ 
  & \textbf{S} & \textbf{M} & \textbf{L} & \textbf{O}   \\ 
 \midrule
 None & 75.4 & 62.6 & 51.8 & 63.3 & 54.4 \\
 \rowcolor{cyan!10} Bilinear Interpolation & \textbf{77.1} & \textbf{64.9} & \textbf{55.7} & \textbf{65.9} & \underline{64.7} \\ 
 Adaptive Pooling & 75.7 & \underline{63.0} & 53.1 & 63.9 & \textbf{64.8} \\
 Dynamic Token Merging & \underline{76.4} & 62.8  & \underline{54.3} & \underline{64.5} & 63.0 \\
  \bottomrule
\end{tabular}
\caption{Results on different token assignment strategies when extending QuoTA with LLaVA-Video-7B \cite{llavavideo} on Video-MME \cite{videomme} and VNBench \cite{vnbench}. ``None'' represents the baseline.}
\label{abl_token}
\end{table}


\noindent \textbf{Effect of different query-oriented frame scoring strategy.} 
As detailed in Sections \ref{subsec:3_2} and \ref{subsec:3_3}, we involve query decoupling into an object list via Chain-of-Thoughts, followed by LVLM-based frame scoring to determine query-relevant importance. 
To evaluate their efficacy, we conducted ablation experiments presented in Table \ref{abl_cot}, indicating that decoupling queries with an emphasis on video event identification (decouple prompt elaborated in Appendix) with slight performance decreases both in two benchmarks, suggesting that entity-based representations constitute fundamental and generalizable features for video understanding.
Notably, when encountering summarization tasks, QuoTA can evenly distribute tokens without forming clusters, as further demonstrated in the Appendix with the sub-tasks evaluation on Video-MME \cite{videomme}.
Furthermore, CLIP \cite{clip} resulted in substantial performance degradation, attributable to CLIP's propensity to prioritize visually anomalous frames (e.g., those with overexposure), consequently compromising accurate query-oriented frame selection.

\begin{table}[]
\centering 
\setlength{\tabcolsep}{1.2mm} 
\renewcommand\arraystretch{0.95} 
\begin{tabular}{l|cccc|c}
\toprule
\multirow{2}{*}{\textbf{Scoring Strategy}} & \multicolumn{4}{c|}{\textbf{Video-MME}} & \multirow{2}{*}{\textbf{VNBench}}   \\ 
  & \textbf{S} & \textbf{M} & \textbf{L} & \textbf{O}   \\ 
 \midrule
 None & 75.4 & 62.6 & 51.8 & 63.3 & 54.4\\
 LVLM-Based & 76.0 & 63.3 & 54.0 & 64.4 & 49.5 \\
 \rowcolor{cyan!10} LVLM-CoT-Entity & \textbf{77.1} & \textbf{64.9} & \underline{55.7} & \textbf{65.9} & \textbf{64.7} \\ 
 LVLM-CoT-Event & \underline{76.2} & \underline{63.9} & \textbf{55.8} & \underline{65.3} &  \underline{64.4} \\
 CLIP-CoT-Entity & 75.8 & 63.6 & 52.8 & 64.0 & 62.7 \\
  \bottomrule
\end{tabular}
\caption{Results on different frame scoring strategies when extending QuoTA with LLaVA-Video-7B \cite{llavavideo} on Video-MME \cite{videomme} and VNBench \cite{vnbench}. ``None'' represents the baseline.}
\label{abl_cot}
\end{table}


\subsection{Qualitative Evaluation}

We present qualitative analyses from Video-MME \cite{videomme} case studies in Figure \ref{fig_exam}, which illustrates normalized frame scores alongside selected sampled frames, with query-relevant keyframes highlighted by blue borders. As demonstrated, QuoTA enables preferential token allocation to query-relevant keyframes while proportionally reducing token assignment to irrelevant frames of LLaVA-Video-7B \cite{llavavideo}, effectively mitigating visual redundancy and facilitating more precise task-specific responses to user queries.
\section{Conclusion}
\label{sec:con}

We present \textbf{QuoTA}, which presents a training-free framework that employs query-oriented visual token assignment for enhanced long video understanding. By leveraging Chain-of-Thoughts reasoning for query-specific frame importance assessment, we achieved a 3.2\% average improvement on LLaVA-Video-7B across six benchmarks while maintaining computational efficiency. The plug-and-play nature of \textbf{QuoTA} ensures seamless integration with existing LVLMs without additional training requirements. Our work demonstrates that query awareness and intelligent token assignment are fundamental to addressing the information redundancy challenges in long video understanding tasks. 
\par\noindent
{\small
    \bibliographystyle{ieeenat_fullname}
    \bibliography{main}
}

\clearpage
\setcounter{page}{1}
\maketitlesupplementary

\section{CoT-Driven Decouple Prompt}

As discussed in the manuscript, we employ a decoupling approach for the user's query using Chain-of-Thoughts (CoT) reasoning \cite{cot} to achieve improved frame-level query-oriented importance scoring. This is accomplished through the use of specially designed decoupled prompts tailored for the language-vision model (LVLM).
We have developed two distinct types of Chain-of-Thoughts reasoning: (i) Object List-Based and (ii) Video Event-Based, with the former demonstrating superior performance.
As illustrated in the accompanying figure, we begin by clearly defining the task, followed by the inclusion of additional in-context examples to effectively prompt the LVLM.
For the CoT-driven decoupling prompt related to the object list, we guide the LVLM to decouple the query by (i) assessing the necessity of entity decoupling, (ii) transforming the original query into a structured object list when appropriate, and (iii) refining this list by eliminating abstract concepts.
In the case of the CoT-driven decoupling prompt for video events, we instruct the LVLM to decouple the query by (i) analyzing the type of the original question (e.g., what, who, how, where, when, why), (ii) identifying the key elements to focus on (such as objects, actions, states, and scenes), and (iii) formulating a simple, direct question to ascertain whether a frame contains these key elements.

\begin{figure}[!htbp]
  \centering
  \includegraphics[width=\linewidth]{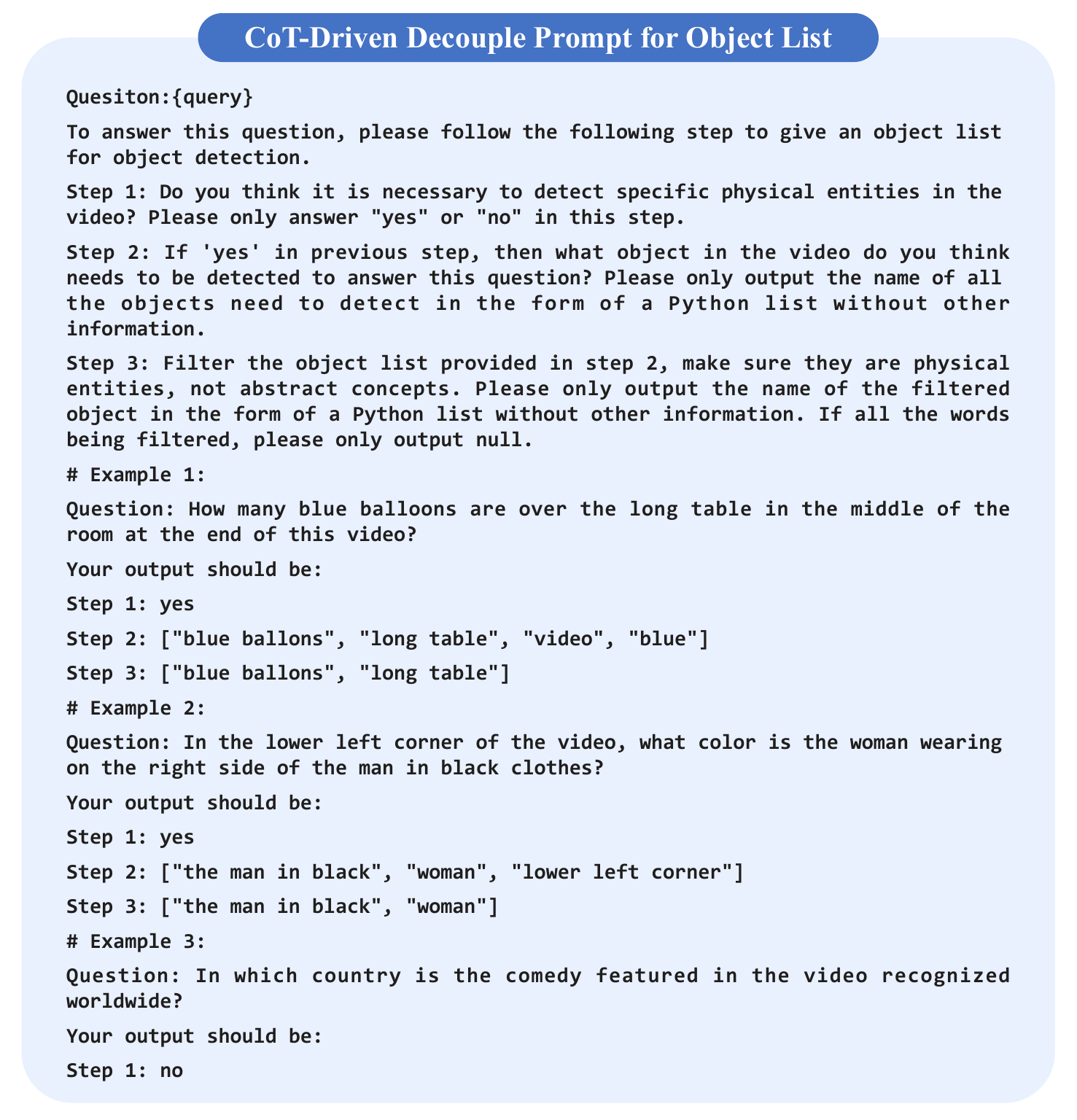}
  \caption{CoT-driven decouple prompt for object list.}
  \label{fig_obj}
\end{figure}

\begin{figure}[!htbp]
  \centering
  \includegraphics[width=\linewidth]{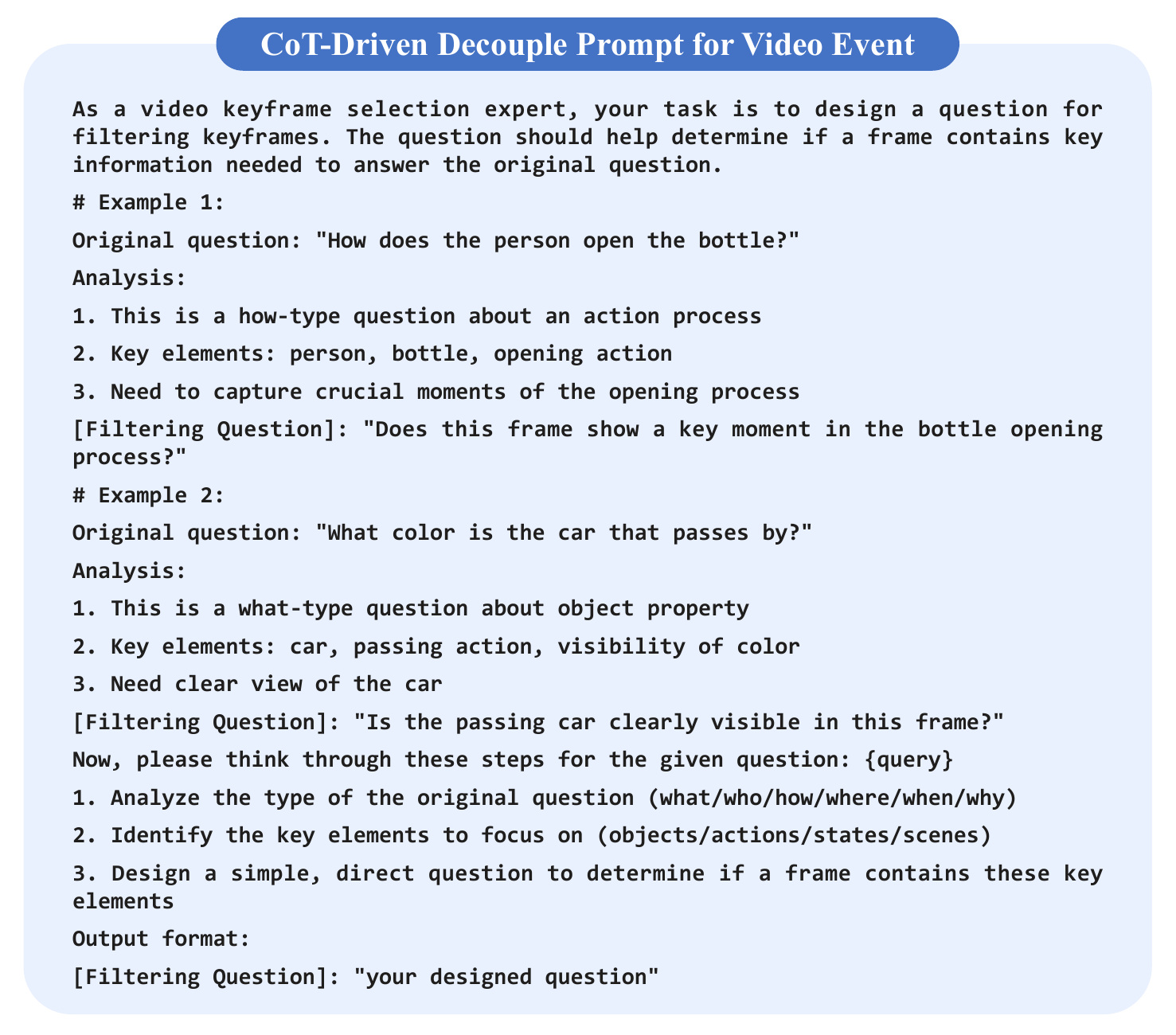}
  \caption{CoT-driven decouple prompt for video event.}
  \label{fig_event}
\end{figure}

\begin{figure*}[h!]
  \centering 
  \includegraphics[width=0.9\linewidth]{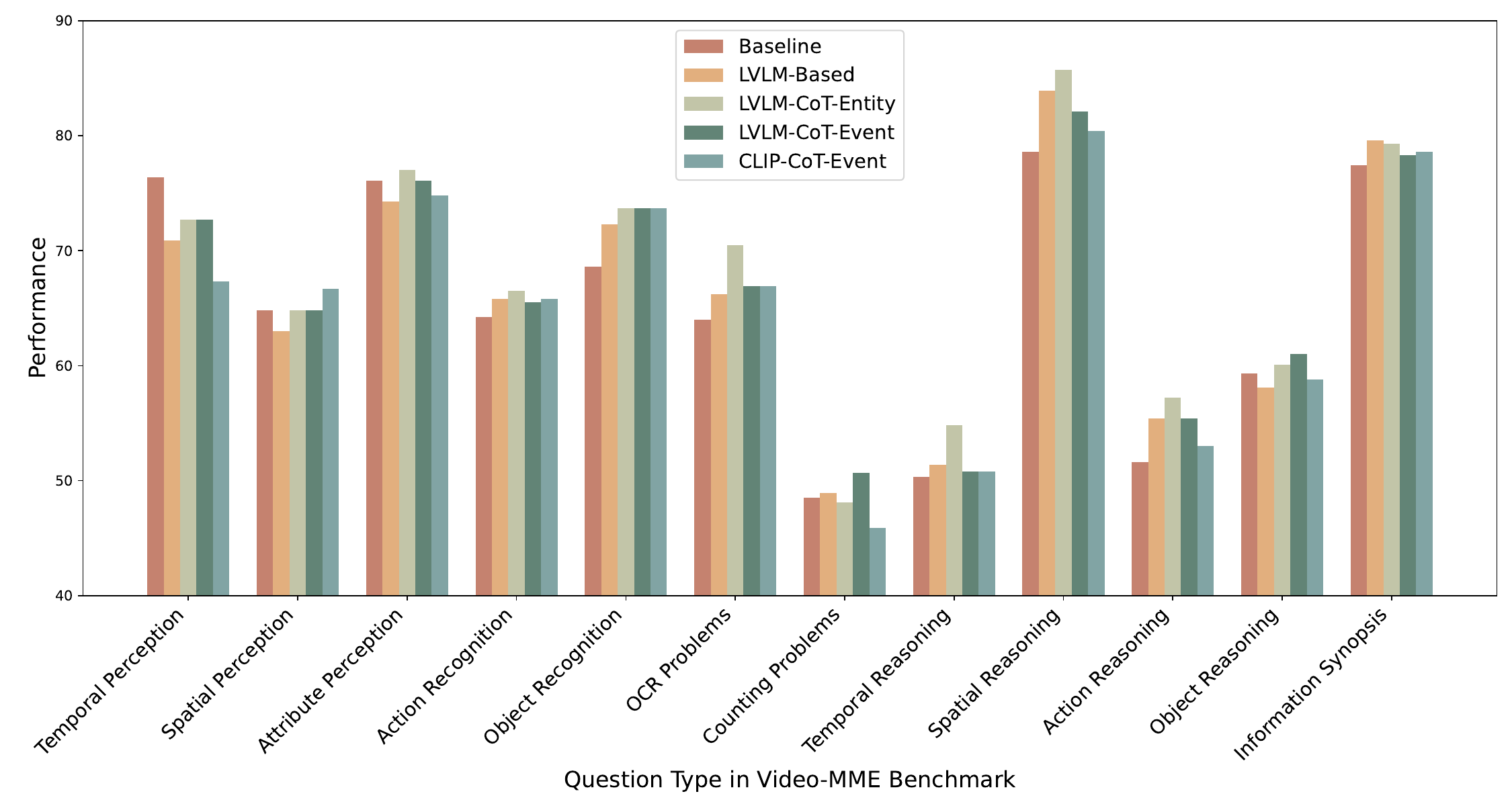}
  \caption{Sub-task results shown in Video-MME \cite{videomme} benchmark when applying distinct frame scoring strategy of LLaVA-Video-7B \cite{llavavideo}.}
  \label{fig_exam} 
\end{figure*}

\section{Effect of Different Frame Scoring Strategy}

We conduct experiments that evaluate the efficacy of query-oriented frame scoring strategies across diverse question types in the Video-MME \cite{videomme} benchmark and VNBench \cite{vnbench}. Figure \ref{fig_exam} reveals insights into our QuoTA underlying different strategies of different sub-tasks in Video-MME \cite{videomme}:

\noindent \textbf{Generalization Advantage of Entity-Centric Decomposition.}
The strategy based on entities demonstrates superior performance in 9 out of 12 question types compared to the baseline, with notable gains in Object Recognition (+5.1\%), OCR Problems (+6.5\%), and Spatial Reasoning (+7.1\%). This aligns with the hypothesis that decomposing queries into entity lists (e.g., "person, bottle, table" for Object Recognition) establishes stable semantic anchors for cross-video generalization. The strategy’s dominance in Information Synopsis (79.3\%) further suggests that entity-based representations mitigate noise from transient visual patterns, enabling robust aggregation of core semantics across temporal spans.

\noindent \textbf{Dynamic Context Sensitivity of Event-Driven Strategies.}
While the strategy based on video events underperforms the entity-based one in most categories, it achieves marginal improvements in Counting Problems (50.7\% vs. 48.1\%) and Object Reasoning (61.0\% vs. 60.1\%). This indicates that event-centric decomposition (e.g., "opening a door → placing an object") better captures transient interactions critical for counting sequential actions. However, its inferior performance in Temporal Reasoning (50.8\%) reveals limitations in modeling long-range dependencies, likely due to fragmented event segmentation that disrupts holistic temporal logic.

\begin{table}[t!]
\centering 
\setlength{\tabcolsep}{1.3mm} 
\renewcommand\arraystretch{0.95} 
\begin{tabular}{l|ccc|c}
\toprule
 \textbf{Scoring Strategy} & \textbf{Count} & \textbf{Order} & \textbf{Retrieve} & \textbf{Overall}   \\ \midrule
 None & 24.0 & 62.7 & 76.4 & 54.4 \\
 LVLM-Based & 18.9 & 74.7 & 54.7 & 49.5 \\
 \rowcolor{cyan!10} LVLM-CoT-Entity & \textbf{29.7} & \textbf{80.4} & 84.0 & \textbf{64.7}  \\ 
 LVLM-CoT-Event & \underline{27.6} & \underline{79.8} & \textbf{85.8} & \underline{64.4} \\
 CLIP-CoT-Entity & 26.7 & 76.0 & \underline{84.7} & 62.4 \\
  \bottomrule
\end{tabular}
\caption{Results on different frame scoring strategies when extending QuoTA with LLaVA-Video-7B \cite{llavavideo} on VNBench \cite{vnbench}.}
\label{abl}
\end{table}

\noindent \textbf{Semantic-Visual Misalignment in CLIP-Based Scoring.}
Frame scoring based on CLIP \cite{clip} exhibits paradoxical behavior: it achieves competitive performance in low-semantic tasks like Spatial Perception (66.7\%) but severely degrades in high-level reasoning tasks (Temporal Reasoning: 50.8\%). This dichotomy stems from CLIP’s inherent bias toward visually salient frames (e.g., motion-blurred or overexposed frames) rather than semantically scenes. For instance, in Attribute Perception, CLIP’s focus on anomalous frames reduces its ability to aggregate consistent attribute features across time, resulting in a 2.7\% drop compared to the entity-based strategy.

Table \ref{abl} also shows distinct frame scoring strategies' performance under counting, ordering, and retrieving sub-tasks on VNBench \cite{vnbench}, which also demonstrated the superior performance of query-oriented frame scoring strategy with CoT-driven query decoupling.

\end{document}